\documentclass[letterpaper]{article} 
\usepackage{aaai2027}

\nocopyright

\usepackage[hyphens]{url}  
\usepackage{graphicx} 
\usepackage{natbib}  
\usepackage{caption} 
\usepackage{algorithm}

\usepackage{newfloat}
\usepackage{listings}
\DeclareCaptionStyle{ruled}{labelfont=normalfont,labelsep=colon,strut=off} 
\floatstyle{ruled}
\newfloat{listing}{tb}{lst}{}
\floatname{listing}{Listing}

\usepackage{booktabs}

\usepackage{latexsym}

\usepackage[T1]{fontenc}

\usepackage[utf8]{inputenc}

\usepackage{microtype}

\usepackage{graphicx}

\usepackage{multirow}
\usepackage{booktabs} 
\usepackage{adjustbox}
\usepackage{colortbl}
\definecolor{color3}{rgb}{0.95,0.95,0.95} 
\usepackage{amsmath}
\usepackage{amsfonts}
\usepackage{algorithm}
\usepackage{algpseudocode}
\usepackage{subcaption}

\usepackage{hyperref}

\title{FOCUS: FP4 Optimization via Coupled-Relaxation and Dual-Granularity Scaling}

\author{
    Xianglong Yan\textsuperscript{\rm 1}\equalcontrib
    \thanks{Work done during an internship at Tencent.},
    Hong Liu\textsuperscript{\rm 2}\equalcontrib,
    Chengzhu Bao\textsuperscript{\rm 1},
    Tianao Zhang\textsuperscript{\rm 1},
    \\
    Guanghua Yu\textsuperscript{\rm 2},
    Jianchen Zhu\textsuperscript{\rm 2},
    Yulun Zhang\textsuperscript{\rm 1}\corresponding
}

\affiliations{
    \textsuperscript{\rm 1}Shanghai Jiao Tong University\\
    \textsuperscript{\rm 2}Tencent\\
    Correspondence:
    \href{mailto:yulun100@gmail.com}{yulun100@gmail.com}
}

\usepackage{fancyhdr}
\usepackage{graphicx}
\usepackage{datetime}  

\fancypagestyle{firstpagestyle}{%
  \fancyhf{}  
  \fancyhead[L]{\includegraphics[height=0.66cm]{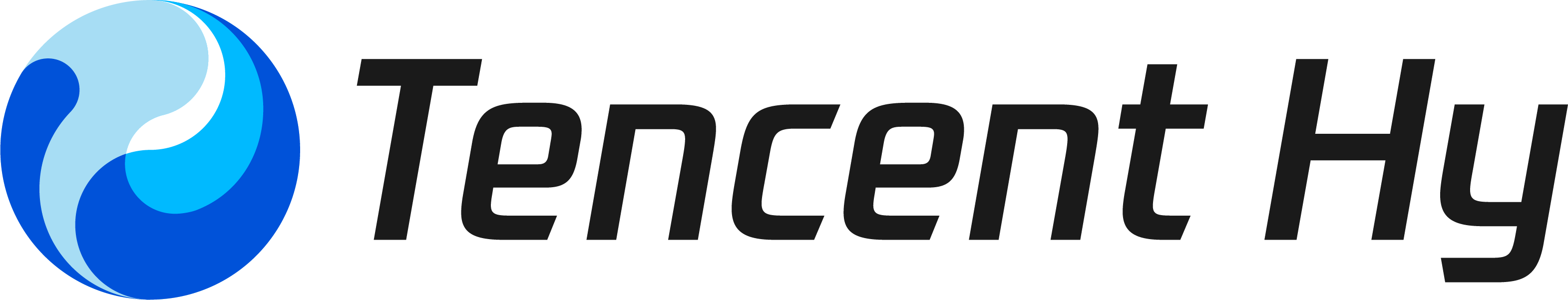}}

}

\begin{document}

\maketitle
\thispagestyle{firstpagestyle}

\begin{abstract}
Large language models (LLMs) achieve remarkable performance but are expensive to deploy due to their enormous size. FP4 quantization, with formats such as MXFP4 and NVFP4, offers an appealing solution with native hardware support on modern accelerators. However, maintaining accuracy under FP4 precision remains difficult. A key bottleneck lies in scale optimization: existing methods tightly couple the quantization and dequantization scales, forcing both to conform to the discrete low-precision format required by hardware (e.g., E8M0 in MXFP4). Yet the quantization scale is never stored and need not obey this constraint, suggesting a significant untapped optimization space. In this work, we propose \textbf{FOCUS}, a post-training quantization framework with end-to-end scale learning for \underline{\textbf{F}}P4 \underline{\textbf{O}}ptimization via \underline{\textbf{C}}oupled-Relaxation and D\underline{\textbf{u}}al-Granularity \underline{\textbf{S}}caling. Coupled-Relaxation Scaling (CRS) relaxes the tight coupling between quantization and dequantization scales with a learnable full-precision coefficient, enabling more effective optimization without breaking hardware compliance. Dual-Granularity Scaling (DGS) further refines the quantization scale at a finer sub-block granularity, allowing more precise adaptation to local weight distributions. Experiments across multiple LLM families and benchmarks show that FOCUS achieves state-of-the-art FP4 accuracy under both MXFP4 and NVFP4 formats, while introducing no additional inference overhead. We will release the code and quantized models of FOCUS at \url{https://github.com/tencent/AngelSlim}.
\end{abstract}

\setlength{\abovedisplayskip}{2pt}
\setlength{\belowdisplayskip}{2pt}


\section{Introduction}
Large language models (LLMs)~\cite{qwen3,llama3} have demonstrated remarkable capabilities across a wide range of tasks. However, modern LLMs~\cite{deepseek,kimi,glm} have scaled to hundreds of billions or even trillions of parameters, posing significant challenges for practical deployment due to prohibitive memory and computational costs.

\begin{figure}[t]
  \centering
  \includegraphics[width=\linewidth]{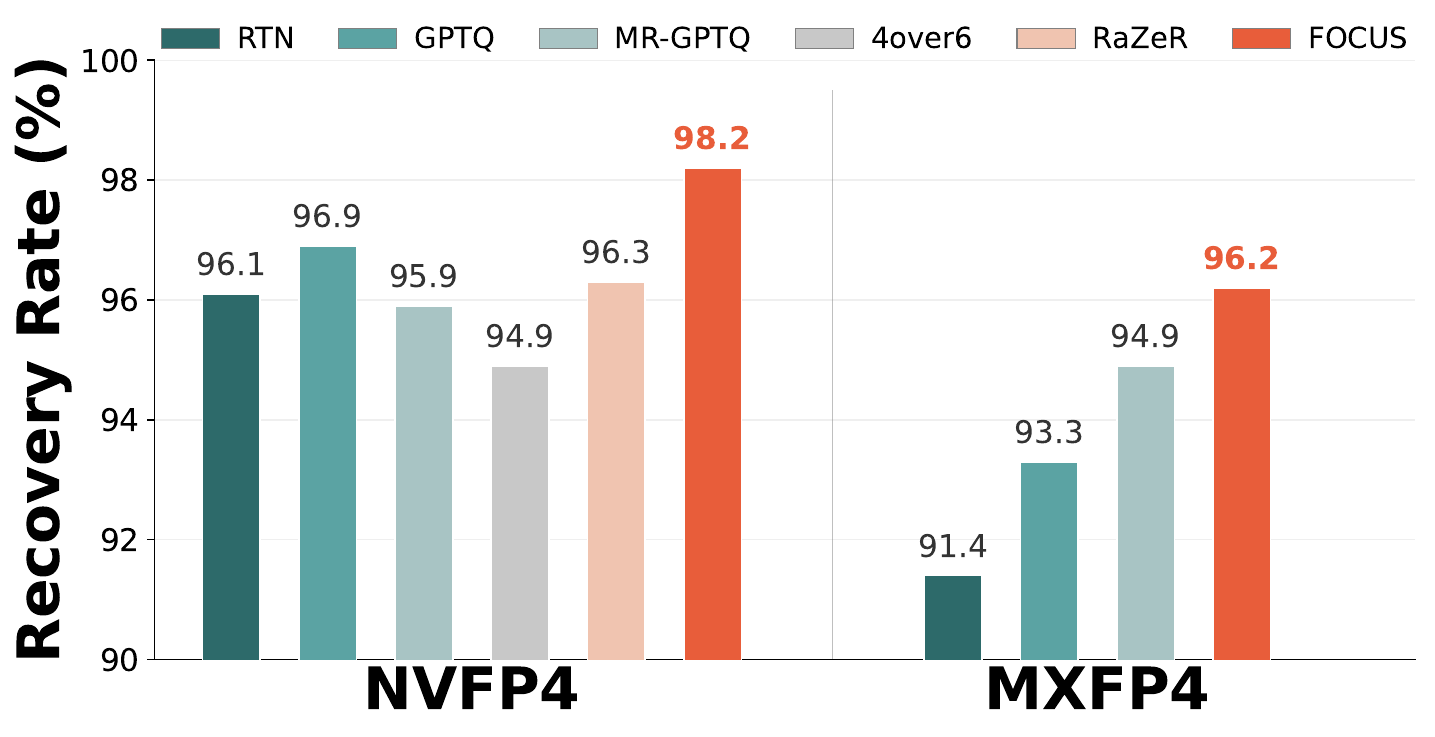}
  \vspace{-1mm} 
    \caption{\textbf{FP4 quantization performance on Qwen3-4B.} Recovery rate is the average zero-shot accuracy relative to FP16. Our method consistently outperforms all baselines under both NVFP4 and MXFP4 formats.}
  \label{figs:fig1}
\vspace{-3mm}
\end{figure}

Post-training quantization (PTQ)~\cite{omniquant,smoothquant,llm-fp4}, which freezes the original model weights and optimizes quantization parameters using only a small calibration set, has become a widely adopted, lightweight, and practical approach for compressing LLMs. As the demand for efficiency grows, quantization has been pushed from 8-bit~\cite{llmint8} to 4-bit~\cite{quarot} and even lower precision~\cite{pt2-llm,arb-llm}. Recently, microscaling floating-point formats such as MXFP4~\cite{mxfp4} and NVFP4~\cite{nvfp4} have emerged with native hardware support on modern accelerators (e.g., NVIDIA Blackwell~\cite{blackwell}), offering a promising path toward efficient W4A4 inference. These formats adopt a block-wise scaling scheme: a group of values (32 for MXFP4, 16 for NVFP4) shares a low-precision scale (E8M0 for MXFP4, E4M3 for NVFP4), enabling compact representation while preserving dynamic range~\cite{intvsfp}.

Despite the hardware support, maintaining model accuracy under FP4 precision remains challenging (see Figure~\ref{figs:fig1}). Recent efforts have explored various directions, including residual compensation~\cite{arcquant}, format-aware rounding~\cite{faar}, redundant zero remapping~\cite{razer}, and transformation-based methods~\cite{mr-gptq,blockrotation,duquant++,batquant}. Among them, transformation-based methods achieve strong performance by reshaping weight and activation distributions, but they introduce online computational overhead (e.g., Hadamard rotation) that requires additional adaptation in inference frameworks. Beyond these, another line of work focuses on scale optimization~\cite{46,unveiling}, yet these approaches rely on simple heuristic rules and yield limited improvements. Overall, scale optimization for FP4 formats remains largely underexplored in current LLM quantization, and no existing method leverages learnable, gradient-based scale optimization.

\begin{figure}[t]
  \centering
  \includegraphics[width=\linewidth]{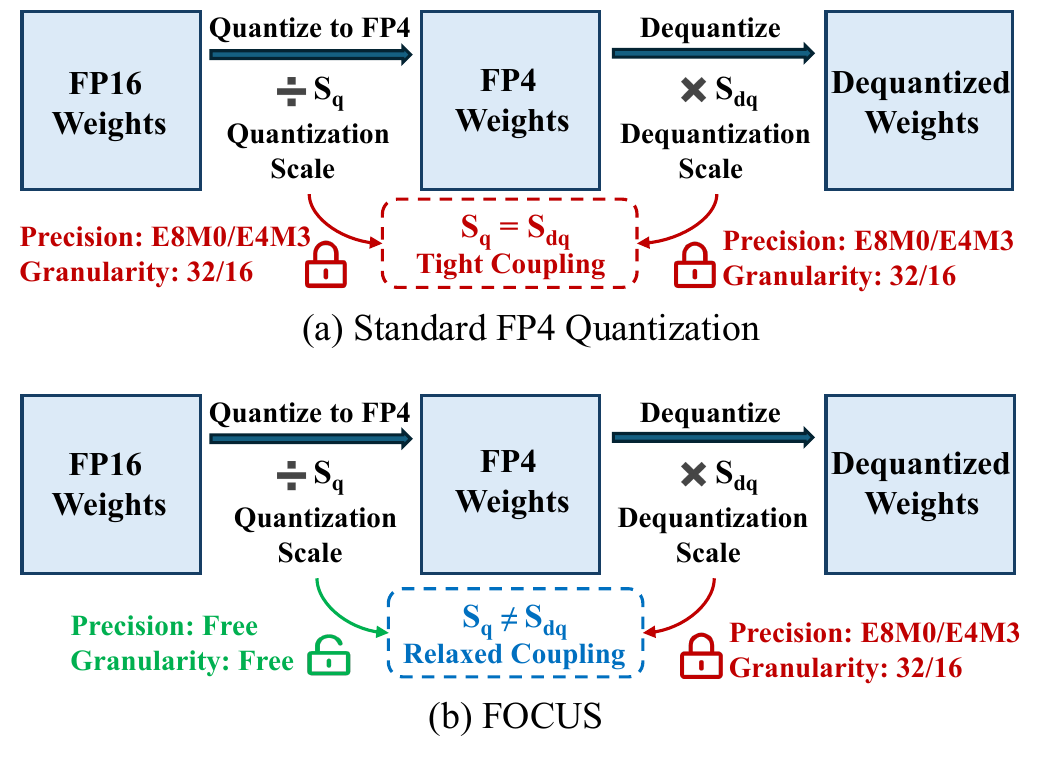}
  \vspace{-6mm} 
    \caption{\textbf{Standard FP4 quantization vs. FOCUS.} (a) Conventional methods tightly couple $S_q$ and $S_{dq}$, constraining both to the same low-precision format and block granularity. (b) FOCUS relaxes this coupling, freeing the quantization scale from the hardware-imposed precision and granularity constraints.}
  \label{figs:fig2}
\vspace{-3mm}
\end{figure}

However, optimizing scales for FP4 formats is inherently difficult~\cite{benchmarkingfp4}. The scale factor itself must be quantized to a low-precision format, restricting optimization to a discrete space. Furthermore, existing methods~\cite{46,unveiling} tightly couple the quantization and dequantization scales into a single shared value, forcing both to bear the precision loss from this hardware constraint. We note that only the dequantization scale is stored and deployed at inference time, and thus must conform to hardware requirements. The quantization scale, by contrast, is used solely during the offline quantization process and is never retained. This leads to a key insight: the quantization scale need not share the same precision format or block-size granularity as the dequantization scale, and can be optimized without any hardware or inference constraint. We illustrate this in Figure~\ref{figs:fig2}.

Building on this insight, we propose \textbf{FOCUS}, a post-training quantization (PTQ) framework with end-to-end scale learning for \underline{\textbf{F}}P4 \underline{\textbf{O}}ptimization via \underline{\textbf{C}}oupled-Relaxation and D\underline{\textbf{u}}al-Granularity \underline{\textbf{S}}caling. FOCUS exploits the freedom of the quantization scale along two dimensions. \textbf{Along the precision dimension}, the quantization scale need not conform to the discrete low-precision format imposed by hardware. Coupled-Relaxation Scaling (CRS) leverages this by introducing a learnable full-precision coefficient that relaxes the tight coupling between quantization and dequantization scales. This lifts the quantization scale optimization from the discrete low-precision space to a continuous full-precision space amenable to gradient-based learning. \textbf{Along the granularity dimension}, the quantization scale likewise need not follow the predefined block size. Dual-Granularity Scaling (DGS) utilizes this by assigning quantization scales at sub-block level. This allows the quantization scale to operate at a finer granularity than the dequantization scale, capturing local weight variations that a block-wise scale cannot represent. 

Extensive experiments across multiple LLM families demonstrate that FOCUS achieves state-of-the-art (SOTA) FP4 accuracy under both MXFP4 and NVFP4 formats. As illustrated in Figure~\ref{figs:fig1}, FOCUS recovers 98.2\% of the FP16 zero-shot accuracy on Qwen3-4B under NVFP4, consistently outperforming all existing baselines. Moreover, since FOCUS only learns scale parameters, quantizing Qwen3-4B takes only 19 minutes on a single H20 GPU with no additional inference overhead. 

\begin{figure*}[t]  
\centering
\includegraphics[width=\textwidth]{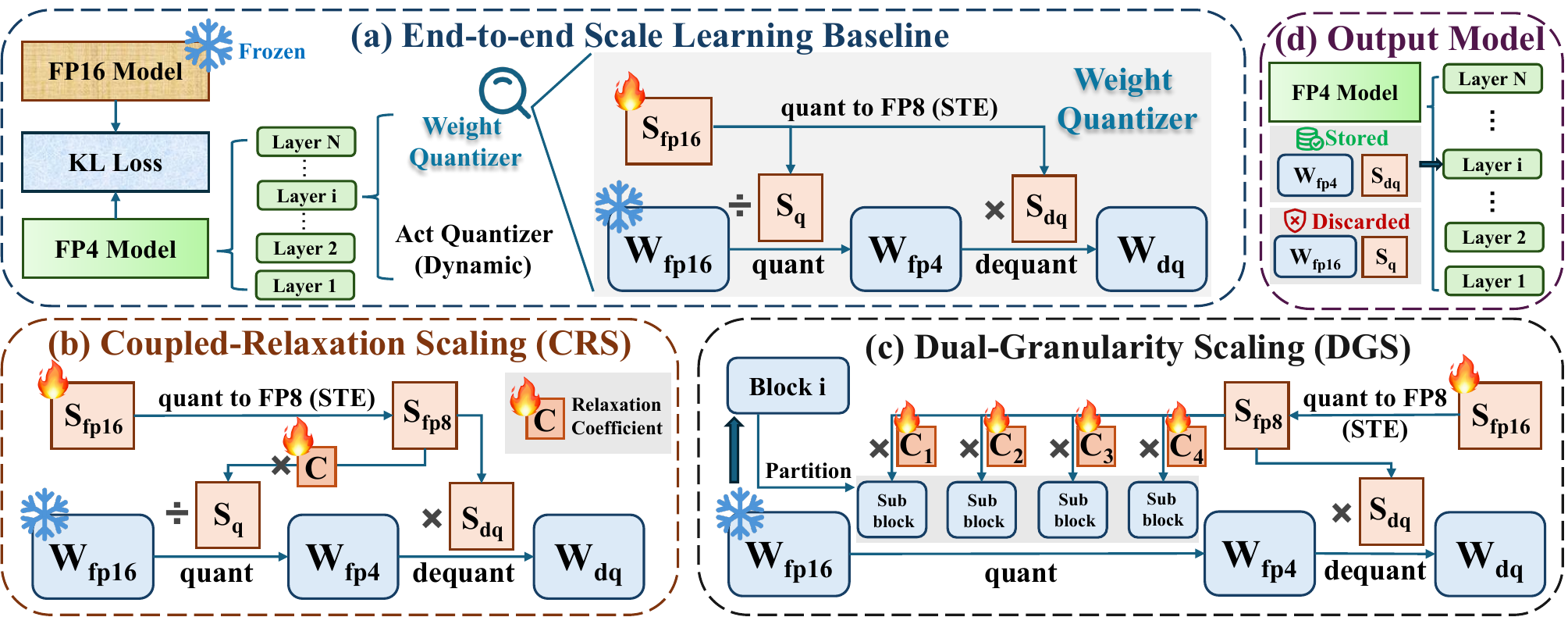}
\caption{Overview of FOCUS. \textbf{(a) End-to-end Scale Learning Baseline}: learns FP8 scales via STE while weights are frozen during quantization and dequantization. \textbf{(b) Coupled-Relaxation Scaling (CRS)}: introduces learnable relaxation coefficients for additional flexibility beyond FP8 precision. \textbf{(c) Dual-Granularity Scaling (DGS)}: partitions each block into sub-blocks with independent coefficients for fine-grained scale adjustment. \textbf{(d) Output Model}: only FP4 weights and dequantization scales are stored, while quantization scales are discarded.}
\label{overview}
\vspace{-3mm}
\label{figs:fig3}
\end{figure*}

Our key contributions are summarized below:
\begin{itemize}
\item We identify that the quantization scale is free from hardware constraints, and propose FOCUS, a PTQ scale learning framework for MXFP4 and NVFP4 that exploits this freedom in both precision and granularity.
\item We propose Coupled-Relaxation Scaling (CRS), which introduces a learnable full-precision coefficient to relax the quantization-dequantization scale coupling, lifting optimization from discrete to continuous space.
\item We propose Dual-Granularity Scaling (DGS), which assigns quantization scales at sub-block level, capturing local weight variations that a single block-wise scale cannot represent.
\item Extensive experiments demonstrate that FOCUS achieves SOTA FP4 accuracy under both MXFP4 and NVFP4, with no inference overhead and seamless framework compatibility.
\end{itemize}

\section{Related Work}
\subsection{PTQ with Learnable Quantization Parameters} 
To improve the accuracy of post-training quantization (PTQ), recent works introduce learnable parameters into the quantization pipeline while keeping the original model weights frozen. Several methods learn quantization scales or clipping bounds: OmniQuant~\cite{omniquant} jointly optimizes learnable weight clipping and equivalent transformations via block-wise distillation, while LRQuant~\cite{lrquant} learns smoothing parameters for enhanced robustness. A complementary direction reshapes distributions through learned transformations. AffineQuant~\cite{affinequant} extends equivalent transformations to full affine mappings. SpinQuant~\cite{spinquant} and OSTQuant~\cite{ostquant} further optimize learned rotation and scaling transformations. FlatQuant~\cite{flatquant} accelerates such optimization while maintaining strong performance. On the rounding side, AdaRound~\cite{adaround} formulates rounding as continuous optimization, which BRECQ~\cite{brecq} extends with block-wise reconstruction. However, these methods are designed for uniform integer grids, where quantization levels are evenly spaced and scales remain in full precision. FP4 formats like MXFP4 and NVFP4 introduce non-uniform grids and low-precision scales (e.g., E8M0), making existing approaches not directly transferable.

\subsection{LLM Quantization with FP4}
Recent FP4 formats such as MXFP4~\cite{mxfp4} and NVFP4~\cite{nvfp4} use block-wise shared scale factors, offering better dynamic range than integer formats. A line of work addresses the incompatibility between global rotations and block-wise scaling by proposing block-wise~\cite{mr-gptq, blockrotation} or outlier-aware~\cite{duquant++} rotation strategies, while BATQuant~\cite{batquant} restricts affine transformations to align with the microscaling granularity. Another line focuses on the scale factor itself: \citet{benchmarkingfp4} identify it as a critical error source, Four Over Six~\cite{46} adaptively selects between scale candidates, \citet{unveiling} propose overflow-aware scaling to improve fidelity, and \citet{soar} optimize NVFP4 scales via closed-form analytical updates with decoupled scale search. Beyond these, ARCQuant~\cite{arcquant} compensates error via residual channels, RaZeR~\cite{razer} exploits redundant zeros to extend representable values, and FAAR~\cite{faar} proposes format-aware adaptive rounding for NVFP4. Despite these efforts, existing methods either introduce additional inference overhead or optimize scales without end-to-end task-level supervision. No prior work directly learns FP4 scales through gradient-based optimization, which we address in this work.

\section{Methodology}

In this section, we present FOCUS as illustrated in Figure~\ref{figs:fig3}. We begin with FP4 preliminaries, then introduce a scale-learning baseline and the key insight behind FOCUS. We next present two core components, CRS and DGS, which relax the precision and granularity constraints on quantization scales. Finally, we discuss deployment considerations without transformation-based overhead.

\subsection{Preliminaries}
\label{sec:prelim}
Both NVFP4 and MXFP4 represent weights using FP4 (E2M1) elements with block-wise scaling. Given a full-precision tensor $\mathbf{X}$ partitioned into blocks $\{\mathbf{X}_i\}$ of size $B$, each block is quantized and dequantized as:
\begin{equation}
\bar{\mathbf{X}}_i = \mathcal{Q}_{\text{E2M1}}\!\left(\mathbf{X}_i \,/\, S_i\right), \quad \hat{\mathbf{X}}_i = \bar{\mathbf{X}}_i \cdot S_i,
\label{eq:fp4_quant}
\end{equation}
where $S_i$ is the per-block effective scale, $\mathcal{Q}_{\text{E2M1}}(\cdot)$ rounds each element to the nearest E2M1 value, $\bar{\mathbf{X}}_i$ is the quantized FP4 weight matrix, and $\hat{\mathbf{X}}_i$ is the dequantized reconstruction.

\textbf{NVFP4} adopts a two-level scaling mechanism with block size $B=16$. A tensor-wide FP32 global scale $\alpha$ is first computed, followed by per-block FP8 (E4M3) scales $\Delta_i$:
\begin{equation}
\alpha = \tfrac{\max(|\mathbf{X}|)}{M_{\text{FP4}} \cdot M_{\text{FP8}}}, \quad \Delta_i = \mathcal{Q}_{\text{E4M3}}\!\left(\tfrac{\max(|\mathbf{X}_i|)}{\alpha \cdot M_{\text{FP4}}}\right),
\label{eq:nvfp4_scale}
\end{equation}
where $M_{\text{FP4}}=6$ and $M_{\text{FP8}}=448$ denote the maximum representable values of E2M1 and E4M3, and the effective per-block scale is $S_i = \alpha \cdot \Delta_i$.

\textbf{MXFP4} follows the OCP Microscaling specification~\cite{mxfp4_ocp} with block size $B=32$. Each block shares a single E8M0 scale factor that encodes only a power-of-two exponent:
\begin{equation}
S_i = \mathcal{Q}_{\text{E8M0}}\!\left(\max(|\mathbf{X}_i|) \,/\, M_{\text{FP4}}\right),
\label{eq:mxfp4_scale}
\end{equation}
where $\mathcal{Q}_{\text{E8M0}}(\cdot)$ rounds the scale to the nearest power of two (typically ceiling in practice). Compared to NVFP4, MXFP4 uses a larger block size and a more restrictive scale format (E8M0 vs.\ E4M3), making its scale factor coarser and optimization more challenging.

\subsection{Baseline Framework and Key Insight}
\label{sec:baseline}
\textbf{Scale Learning Baseline.}
Unlike coarse per-channel or per-tensor scales in integer quantization, the block-wise scales in NVFP4 and MXFP4 operate at a fine granularity (block size 16 and 32, respectively), making their quality critically important to quantization accuracy. The default absmax initialization (Eqs.~\ref{eq:nvfp4_scale}--\ref{eq:mxfp4_scale}) often yields suboptimal scale values. Recent works have proposed improvements: MR-GPTQ~\cite{mr-gptq} applies MSE-based scale search, while some works~\cite{46,unveiling} adjust the maximum-value mapping rule. However, these methods remain rule-based and operate only on a per-layer basis without leveraging global task loss information, still incurring notable accuracy loss. Inspired by end-to-end quantization-aware optimization~\cite{spinquant,ostquant}, we construct a baseline that optimizes $S_i$ end-to-end.

We freeze the original model weights and optimize only the weight block scales, while keeping activation quantization fully dynamic. For NVFP4, we further freeze the tensor-wide scale $\alpha$ and learn only $\Delta_i$, while for MXFP4, $S_i$ is the sole block-wise scale. Each learnable scale is maintained as a full-precision (FP16) copy $S_i^{\text{fp}}$ for gradient-based optimization, initialized via absmax. During the forward pass, it is quantized on-the-fly to its hardware-mandated format before being applied:
\begin{equation}
S_i = \mathcal{Q}_{\text{fmt}}\!\left(S_i^{\text{fp}}\right),
\label{eq:scale_quant}
\end{equation}
where $\mathcal{Q}_{\text{fmt}}$ denotes quantization to the target scale format (E4M3 or E8M0). The quantized $S_i$ is then used in Eq.~\ref{eq:fp4_quant} for weight quantization and dequantization. The Straight-Through Estimator (STE)~\cite{ste} is applied to all non-differentiable operations, allowing gradients to flow through to $S_i^{\text{fp}}$. Following~\citet{ostquant}, we use a knowledge distillation loss for end-to-end scale optimization.

\textbf{Key Insight.}
In the baseline above, the same quantized scale $S_i$ serves both as the quantization scale (dividing the weights to obtain FP4 values) and the dequantization scale (multiplying to reconstruct the weights). To make this distinction explicit, we rewrite the process as:
\begin{equation}
\bar{\mathbf{W}}_i = \mathcal{Q}_{\text{E2M1}}\!\left(\mathbf{W}_i \,/\, S_i^{\text{q}}\right), \quad \hat{\mathbf{W}}_i = \bar{\mathbf{W}}_i \cdot S_i^{\text{dq}},
\label{eq:qdq}
\end{equation}
where the baseline enforces $S_i^{\text{q}} = S_i^{\text{dq}} = \mathcal{Q}_{\text{fmt}}(S_i^{\text{fp}})$, i.e., the two scales are tightly coupled.

However, these two scales serve fundamentally different roles. The dequantization scale $S_i^{\text{dq}}$ must be stored and deployed at inference time, and therefore must conform to the hardware-mandated format (E4M3 or E8M0). The quantization scale $S_i^{\text{q}}$, by contrast, is used only during the offline quantization process and is never retained. Its role is solely to determine the FP4 weight matrix $\bar{\mathbf{W}}_i$, i.e., to find the best possible FP4 assignment for each block. Since $S_i^{\text{q}}$ is not subject to any hardware or inference constraint, its optimization space can be significantly enlarged to yield better FP4 matrices.

Specifically, the baseline unnecessarily restricts $S_i^{\text{q}}$ along two dimensions:
\begin{itemize}
\item \textbf{Precision.} $S_i^{\text{q}}$ is forced into the same discrete low-precision format as $S_i^{\text{dq}}$ (e.g., E8M0, encoding only power-of-two values), severely limiting the optimization landscape.
\item \textbf{Granularity.} $S_i^{\text{q}}$ is forced to share the same block size as $S_i^{\text{dq}}$ (16 or 32 elements), unable to adapt to intra-block weight variation.
\end{itemize}
These observations motivate our two techniques: Coupled-Relaxation Scaling releases the precision constraint on $S_i^{\text{q}}$, and Dual-Granularity Scaling releases the granularity constraint. Both aim to enlarge the optimization space of $S_i^{\text{q}}$ for finding better FP4 weight matrices, while keeping $S_i^{\text{dq}}$ fully hardware-compliant.

\subsection{Coupled-Relaxation Scaling}
\label{sec:crs}
As identified above, the quantization scale $S_i^{\text{q}}$ is never stored or used at inference time. Only the FP4 matrix $\bar{\mathbf{W}}_i$ and the dequantization scale $S_i^{\text{dq}}$ are retained. Therefore, $S_i^{\text{q}}$ need not obey the hardware precision constraint. A natural approach is to fully decouple the two scales: let $S_i^{\text{q}}$ be a free full-precision parameter independent of $S_i^{\text{dq}}$. However, full decoupling leads to training instability, as the two scales receive conflicting gradients and cause optimization to oscillate (see Appendix).

We instead propose Coupled-Relaxation Scaling (CRS), which retains the coupling structure but relaxes its tightness. Specifically, we introduce a learnable full-precision coefficient $c_i$ (initialized to 1) and define the quantization scale as:
\begin{equation}
S_i^{\text{q}} = S_i^{\text{dq}} \cdot c_i, \quad \text{where} \quad S_i^{\text{dq}} = \mathcal{Q}_{\text{fmt}}(S_i^{\text{fp}}).
\label{eq:crs_scale}
\end{equation}
The quantization and dequantization are then:
\begin{equation}
\bar{\mathbf{W}}_i = \mathcal{Q}_{\text{E2M1}}\!\left(\mathbf{W}_i \,/\, S_i^{\text{q}}\right), \quad \hat{\mathbf{W}}_i = \bar{\mathbf{W}}_i \cdot S_i^{\text{dq}}.
\label{eq:crs}
\end{equation}
Since $c_i$ is continuous and anchored to $S_i^{\text{dq}}$, the optimization of the quantization scale is lifted from the discrete hardware format to a continuous space, while the shared structure ensures training stability.

\begin{figure}[t]
  \centering
  \includegraphics[width=\linewidth]{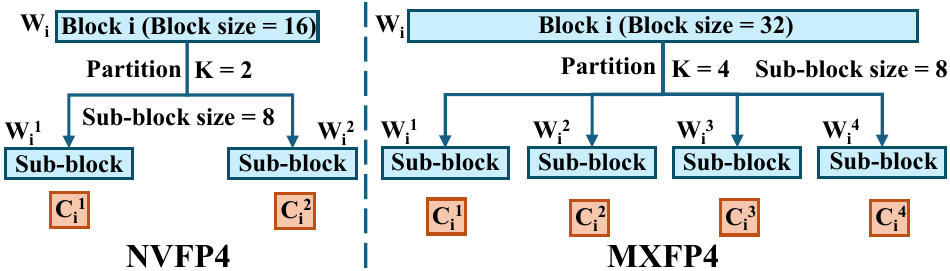}
\caption{Sub-block partition in Dual-Granularity Scaling for NVFP4 and MXFP4.}  
\label{fig:dgs_partition}
\vspace{-3mm}
\end{figure}

We note that CRS introduces an asymmetry: the quantization and dequantization scales are no longer mathematically equivalent. However, the goal of FP4 quantization is not to preserve this equivalence, but to find the FP4 matrix $\bar{\mathbf{W}}_i$ and dequantization scale $S_i^{\text{dq}}$ that maximize end-to-end model performance. Starting from $c_i = 1$ (equivalent to the baseline), the coefficient is driven by the task loss to discover better FP4 assignments.

\begin{table*}[t]
\setlength{\tabcolsep}{5pt}
\small
\centering
\vspace{-3mm}
\begin{tabular}{c|c|l|cc|ccccc|c}
\hline
\textbf{Model} & \textbf{Format} & \textbf{Method} & \textbf{Wiki2$(\downarrow)$} & \textbf{C4$(\downarrow)$} & \textbf{ARC-C} & \textbf{ARC-E} & \textbf{Hella.} & \textbf{PiQA} & \textbf{Winog.} & \textbf{Avg.$(\uparrow)$} \\
\hline
\multirow{12}{*}{\shortstack{Qwen3\\4B}} 
& FP16 & FP16 & 13.66 & 16.63 & 54.18 & 78.07 & 68.50 & 74.81 & 65.67 & 68.25 \\
\cline{2-11}
& \multirow{4}{*}{MXFP4} 
& RTN & 18.60 & 20.62 & 46.08 & 70.41 & 62.91 & 71.60 & 60.77 & 62.35 \\
& & GPTQ & 16.85 & 19.37 & 47.95 & 72.81 & 63.09 & 72.91 & 61.48 & 63.65 \\
& & MR-GPTQ & \underline{15.48} & \underline{18.38} & \underline{48.89} & \underline{74.37} & \underline{63.39} & \underline{73.23} & \underline{64.09} & \underline{64.79} \\
& & FOCUS & \textbf{12.85} & \textbf{17.77} & \textbf{49.15} & \textbf{74.83} & \textbf{66.42} & \textbf{73.61} & \textbf{64.25} & \textbf{65.65} \\
\cline{2-11}
& \multirow{6}{*}{NVFP4} 
& RTN & \underline{13.88} & 17.29 & \underline{51.54} & 74.33 & 65.72 & 74.27 & 62.04 & 65.58 \\
& & GPTQ & 13.91 & 17.30 & 51.28 & \underline{75.46} & 66.59 & \underline{74.59} & 62.83 & \underline{66.15} \\
& & MR-GPTQ & 14.64 & 17.60 & 49.57 & 75.34 & 65.86 & 72.91 & \underline{63.61} & 65.46 \\
& & 4over6 & 14.17 & 17.28 & 48.63 & 74.58 & 66.39 & 71.93 & 62.19 & 64.74 \\
& & RaZeR & 14.11 & \underline{17.26} & 50.34 & 74.75 & \underline{67.10} & 73.29 & 63.06 & 65.71 \\
& & FOCUS & \textbf{12.57} & \textbf{16.97} & \textbf{52.56} & \textbf{76.01} & \textbf{67.26} & \textbf{75.03} & \textbf{64.40} & \textbf{67.05} \\
\hline
\multirow{12}{*}{\shortstack{Qwen3\\8B}} 
& FP16 & FP16 & 9.73 & 13.30 & 56.14 & 80.72 & 74.98 & 77.91 & 67.96 & 71.54 \\
\cline{2-11}
& \multirow{4}{*}{MXFP4} 
& RTN & 11.77 & 15.77 & 50.34 & 71.13 & 69.63 & 73.29 & 63.06 & 65.49 \\
& & GPTQ & 11.64 & 15.48 & 50.77 & 74.92 & \underline{69.79} & 75.14 & 66.61 & 67.45 \\
& & MR-GPTQ & \underline{11.45} & \underline{15.41} & \underline{53.24} & \textbf{76.68} & 69.66 & \underline{75.35} & \underline{67.01} & \underline{68.39} \\
& & FOCUS & \textbf{9.76} & \textbf{14.28} & \textbf{53.58} & \underline{75.76} & \textbf{73.47} & \textbf{76.50} & \textbf{67.64} & \textbf{69.39} \\
\cline{2-11}
& \multirow{6}{*}{NVFP4} 
& RTN & 10.04 & 13.77 & \underline{54.78} & 79.29 & 73.14 & \underline{77.31} & 66.46 & 70.20 \\
& & GPTQ & 10.12 & 13.71 & 54.01 & \underline{79.42} & 72.90 & 76.33 & 68.67 & 70.27 \\
& & MR-GPTQ & 10.28 & 13.87 & 53.75 & 79.25 & 72.95 & \textbf{77.64} & 67.01 & 70.12 \\
& & 4over6 & 10.00 & 13.72 & \underline{54.78} & 78.28 & 73.13 & 75.52 & 66.93 & 69.73 \\
& & RaZeR & \underline{9.99} & \underline{13.64} & 54.69 & 79.21 & \underline{73.51} & 76.39 & \underline{68.90} & \underline{70.54} \\
& & FOCUS & \textbf{9.47} & \textbf{13.62} & \textbf{56.14} & \textbf{80.05} & \textbf{73.67} & \underline{77.31} & \textbf{69.69} & \textbf{71.37} \\
\hline
\end{tabular}
\caption{
\textbf{FP4 quantization results on Qwen3.} Perplexity on WikiText2 and C4, and accuracy (\%) on five zero-shot tasks. Best and second-best results (excluding FP16) within each format are marked in \textbf{bold} and \underline{underlined}.
}
\label{tab:main_results}
\vspace{-3mm}
\end{table*}

\subsection{Dual-Granularity Scaling}
\label{sec:dgs}
CRS relaxes the precision constraint on the quantization scale $S_i^{\text{q}}$ via a per-block coefficient $c_i$. We further relax the granularity constraint: since $S_i^{\text{q}}$ is not retained at inference, it need not share the same block size as $S_i^{\text{dq}}$ and can operate at a finer granularity to further increase optimization freedom.

Specifically, we subdivide each hardware block (size 32 for MXFP4, 16 for NVFP4) into $K$ equal sub-blocks, each assigned an independent full-precision coefficient $c_i^k$. Only the quantization scale operates at finer granularity via the same multiplicative form as CRS, while the dequantization scale $S_i^{\text{dq}}$ remains shared across the entire block:
\begin{equation}
\bar{\mathbf{W}}_i^k = \mathcal{Q}_{\text{E2M1}}\!\left(\tfrac{\mathbf{W}_i^k}{S_i^{\text{dq}} \cdot c_i^k}\right), \quad \hat{\mathbf{W}}_i^k = \bar{\mathbf{W}}_i^k \cdot S_i^{\text{dq}},
\label{eq:dgs}
\end{equation}
where $\mathbf{W}_i^k$ denotes the $k$-th sub-block within block $i$, and each $c_i^k$ is initialized to 1. We set the sub-block size to 8 for both formats, yielding $K{=}4$ for MXFP4 and $K{=}2$ for NVFP4 (see Figure~\ref{fig:dgs_partition}). When $K{=}1$, DGS reduces to CRS. By allowing each sub-block to independently adjust its rounding boundary, DGS captures intra-block weight variation that a single block-level coefficient cannot. As with CRS, all coefficients are discarded after training, and the deployed model remains in the standard hardware format with no inference overhead. The additional training cost is negligible, as FOCUS only learns lightweight scale parameters without modifying the weights (see Appendix).

\begin{table*}[t]
\setlength{\tabcolsep}{5pt}
\small
\centering

\vspace{-3mm}
\begin{tabular}{c|c|l|cc|ccccc|c}
\hline
\textbf{Model} & \textbf{Format} & \textbf{Method} & \textbf{Wiki2$(\downarrow)$} & \textbf{C4$(\downarrow)$} & \textbf{ARC-C} & \textbf{ARC-E} & \textbf{Hella.} & \textbf{PiQA} & \textbf{Winog.} & \textbf{Avg.$(\uparrow)$} \\
\hline
\multirow{11}{*}{\shortstack{LLaMA-3.2\\1B-Instruct}} 
& FP16 & FP16 & 13.16 & 18.52 & 37.54 & 63.68 & 61.56 & 75.14 & 61.72 & 59.93 \\
\cline{2-11}
& \multirow{4}{*}{MXFP4} 
& RTN & 21.06 & 28.10 & 32.34 & 57.62 & 52.37 & 69.59 & \underline{57.38} & 53.86 \\
& & GPTQ & 20.75 & 26.70 & 33.02 & 56.44 & 53.89 & 69.91 & 56.83 & 54.02 \\
& & MR-GPTQ & \underline{17.00} & \underline{22.96} & \underline{34.47} & \underline{59.85} & \underline{55.13} & \textbf{71.65} & \textbf{58.33} & \underline{55.89} \\
& & FOCUS & \textbf{16.46} & \textbf{22.87} & \textbf{36.60} & \textbf{60.60} & \textbf{56.12} & \underline{71.55} & 57.14 & \textbf{56.40} \\
\cline{2-11}
& \multirow{6}{*}{NVFP4} 
& RTN & 15.52 & 21.50 & 35.24 & 60.40 & 57.81 & 72.52 & \textbf{60.85} & 57.36 \\
& & GPTQ & 15.15 & 20.84 & \underline{36.43} & \underline{60.86} & 58.51 & \underline{72.80} & \underline{60.30} & \underline{57.78} \\
& & MR-GPTQ & 15.19 & \textbf{20.59} & 34.98 & 60.82 & 57.82 & 72.69 & 59.98 & 57.26 \\
& & 4over6 & 15.34 & 21.53 & 35.92 & 59.72 & 57.91 & 72.52 & \textbf{60.85} & 57.38 \\
& & RaZeR & \textbf{14.79} & 20.79 & 35.75 & 60.35 & \underline{58.87} & 72.25 & 59.04 & 57.25 \\
& & FOCUS & \underline{14.83} & \underline{20.69} & \textbf{36.52} & \textbf{61.15} & \textbf{59.47} & \textbf{73.12} & 60.06 & \textbf{58.06} \\
\hline
\multirow{11}{*}{\shortstack{LLaMA-3.2\\3B-Instruct}} 
& FP16 & FP16 & 11.04 & 14.48 & 45.99 & 71.30 & 71.59 & 76.99 & 68.90 & 66.95 \\
\cline{2-11}
& \multirow{4}{*}{MXFP4} 
& RTN & 15.15 & 18.98 & 42.49 & 65.99 & 65.90 & 73.61 & 64.72 & 62.54 \\
& & GPTQ & 14.72 & 18.08 & 43.34 & 66.71 & 66.23 & 74.86 & \textbf{66.61} & 63.55 \\
& & MR-GPTQ & \underline{13.29} & \textbf{16.56} & \underline{43.43} & \textbf{70.66} & \underline{67.72} & \underline{76.06} & \underline{65.19} & \underline{64.61} \\
& & FOCUS & \textbf{12.30} & \underline{16.64} & \textbf{45.14} & \underline{69.44} & \textbf{69.21} & \textbf{76.12} & 65.11 & \textbf{65.00} \\
\cline{2-11}
& \multirow{6}{*}{NVFP4} 
& RTN & 11.91 & \underline{15.49} & \underline{44.97} & \textbf{70.58} & 70.04 & 75.79 & 66.06 & 65.49 \\
& & GPTQ & 12.16 & 15.54 & 43.43 & 68.39 & 70.37 & 75.95 & 66.85 & 65.00 \\
& & MR-GPTQ & 12.09 & 15.63 & \textbf{45.31} & 70.03 & 69.53 & \textbf{77.09} & \underline{67.96} & \underline{65.98} \\
& & 4over6 & 11.95 & 15.51 & 44.37 & 69.57 & 70.00 & 75.90 & 67.40 & 65.45 \\
& & RaZeR & \underline{11.65} & \textbf{15.28} & 44.88 & 68.14 & \underline{70.82} & 76.12 & 67.40 & 65.47 \\
& & FOCUS & \textbf{11.48} & 15.59 & 44.54 & \underline{70.29} & \textbf{70.91} & \underline{76.39} & \textbf{68.75} & \textbf{66.18} \\
\hline
\multirow{11}{*}{\shortstack{LLaMA-3.1\\8B-Instruct}} 
& FP16 & FP16 & 7.22 & 10.39 & 55.63 & 79.84 & 79.54 & 81.50 & 73.48 & 74.00 \\
\cline{2-11}
& \multirow{4}{*}{MXFP4} 
& RTN & 9.59 & 13.49 & 47.87 & 74.79 & 74.99 & 77.31 & 68.35 & 68.66 \\
& & GPTQ & 8.99 & 12.75 & 49.66 & 74.62 & \underline{76.05} & \underline{79.05} & 71.35 & 70.15 \\
& & MR-GPTQ & \textbf{8.37} & \textbf{12.13} & \underline{51.71} & \underline{76.89} & 75.53 & \underline{79.05} & \underline{72.22} & \underline{71.08} \\
& & FOCUS & \underline{8.40} & \underline{12.17} & \textbf{52.30} & \textbf{78.93} & \textbf{76.75} & \textbf{79.11} & \textbf{72.30} & \textbf{71.88} \\
\cline{2-11}
& \multirow{6}{*}{NVFP4} 
& RTN & 7.89 & 11.37 & 52.65 & 77.02 & 77.95 & \underline{80.14} & 72.22 & 72.00 \\
& & GPTQ & \underline{7.90} & 11.39 & \textbf{54.10} & 78.16 & 77.72 & 78.56 & 72.22 & 72.15 \\
& & MR-GPTQ & 7.97 & 11.41 & 53.33 & \textbf{78.66} & 77.50 & 80.09 & 71.43 & \underline{72.20} \\
& & 4over6 & 7.92 & \underline{11.36} & \underline{53.58} & 76.68 & 78.01 & 79.05 & \underline{72.30} & 71.92 \\
& & RaZeR & 7.98 & \textbf{11.32} & 53.24 & 77.23 & \underline{78.15} & 79.54 & 71.98 & 72.03 \\
& & FOCUS & \textbf{7.89} & \textbf{11.32} & 52.73 & \underline{77.74} & \textbf{78.90} & \textbf{80.36} & \textbf{72.53} & \textbf{72.45} \\
\hline
\end{tabular}
\caption{
\textbf{FP4 quantization results on LLaMA3.} Perplexity on WikiText2 and C4, and accuracy (\%) on five zero-shot tasks. Best and second-best results (excluding FP16) within each format are marked in \textbf{bold} and \underline{underlined}.
}
\label{tab:full_results_llama}
\vspace{-5mm}
\end{table*}

\subsection{FOCUS without Extra Transforms}
\label{sec:no_transform}
Several recent SOTA FP4 methods~\cite{blockrotation,mr-gptq} apply Hadamard transforms to smooth distributions before quantization. FOCUS does not rely on such transforms, and we discuss the relationship below.

\textbf{Practical issues of Hadamard transforms.}
While theoretically fusible into adjacent layers, existing methods~\cite{mr-gptq} do not actually fuse the Hadamard transform in their implementations. We conduct fusion experiments and find that fusing the transform and re-quantizing degrades accuracy compared to online computation (see Appendix). Without fusion, every linear layer incurs additional Hadamard overhead at inference, making the model incompatible with mainstream inference frameworks.

\textbf{Orthogonality with FOCUS.}
FOCUS operates entirely within the scale optimization space and introduces no structural modification to the model. The deployed model remains in the standard MXFP4/NVFP4 format, fully compatible with hardware-native execution. Moreover, FOCUS is complementary to transform-based methods and can be combined with them (see Appendix).

\section{Experiments}

\subsection{Experimental Settings}
\label{sec:experiments_settings}
\textbf{Implementation Details.}
All experiments are conducted using PyTorch~\citep{pytorch} and HuggingFace Transformers~\citep{huggingface} on a single NVIDIA H20 GPU. Following~\citet{ostquant}, we adopt the KL-Top loss with $k=1000$ as the training objective. We quantize both weights and activations of all linear layers to FP4. We use AdamW~\cite{adamw} as the optimizer with a constant learning rate schedule. For MXFP4, the learning rates for the block-wise scale and the relaxation coefficient are set to $2\times10^{-2}$ and $5\times10^{-2}$, respectively. For NVFP4, the corresponding learning rates are $5\times10^{-3}$ and $1\times10^{-3}$. We train for 1 epoch with a global batch size of 32, using 1,248 samples from the WikiText2~\citep{merity2016pointer} training set with a sequence length of 2,048. For DGS, each MXFP4 block (32 elements) is partitioned into 4 sub-blocks of 8 elements, and each NVFP4 block (16 elements) is partitioned into 2 sub-blocks of 8 elements (ablated in Table~\ref{tab:dgs_granularity}).

\textbf{Models and Evaluation.}
We evaluate FOCUS on five representative LLMs spanning different model families and sizes: LLaMA-3.1-8B-Instruct, LLaMA-3.2-1B-Instruct, LLaMA-3.2-3B-Instruct~\citep{llama3}, Qwen3-4B, and Qwen3-8B~\citep{qwen3}. For evaluation, we report perplexity on WikiText2~\citep{merity2016pointer} and C4~\citep{raffel2020exploring} with a sequence length of 2,048 tokens, and assess zero-shot accuracy on five common-sense reasoning benchmarks: ARC-Challenge~\citep{clark2018think}, ARC-Easy~\citep{clark2018think}, HellaSwag~\citep{zellers2019hellaswag}, PIQA~\citep{bisk2020piqa}, and Winogrande~\citep{sakaguchi2019adversarial}. Furthermore, to evaluate the reasoning capabilities of quantized models, we conduct chain-of-thought (CoT) evaluations on GSM8K~\citep{gsm8k} and MMLU~\citep{mmlu}.

\textbf{Baselines.}
We compare FOCUS against both classic and recent PTQ methods under the two FP4 formats. For both MXFP4 and NVFP4, we include RTN (round-to-nearest), GPTQ~\cite{gptq}, and MR-GPTQ~\cite{mr-gptq} as common baselines. For NVFP4, we additionally compare with two format-specific methods: 4over6~\cite{46} and RaZeR~\cite{razer}, the current SOTA under this format.

\subsection{Main Results}
\textbf{Language Modeling (Perplexity).}
As shown in Tables~\ref{tab:main_results} and~\ref{tab:full_results_llama}, FOCUS achieves the lowest perplexity in nearly all settings across five models. Notably, on Qwen3-8B, FOCUS attains 9.76 (MXFP4) and 9.47 (NVFP4) on WikiText2, nearly matching the FP16 baseline of 9.73. The gains are particularly significant under MXFP4, where FOCUS reduces perplexity by 1.5--2.5 points over MR-GPTQ on Qwen models, demonstrating the effectiveness of scale optimization under more restrictive quantization constraints. Under NVFP4, FOCUS remains best or competitive, outperforming format-specific methods such as RaZeR and 4over6 by clear margins.

\begin{table}[h]
\small
\centering
\begin{tabular}{c|l|cc}
\hline
\textbf{Model} & \textbf{Method} & \textbf{GSM8K-CoT} & \textbf{MMLU-CoT}\\
\hline
\multirow{5}{*}{\shortstack{LLaMA-3.2\\3B-Instruct}}
& FP16    & 77.86 & 64.57 \\
\cline{2-4}
& RTN     & 55.42 & 53.96\\
& GPTQ    & 51.10    & 51.60   \\
& MR-GPTQ & 60.20    & 55.16    \\
& FOCUS   & \textbf{64.14} & \textbf{59.09} \\
\hline
\multirow{5}{*}{\shortstack{LLaMA-3.1\\8B-Instruct}}
& FP16    & 85.67 & 72.71 \\
\cline{2-4}
& RTN     & 59.29 & 57.90\\
& GPTQ    & 63.84    & 61.81   \\
& MR-GPTQ & 72.40    & 63.39    \\
& FOCUS   & \textbf{73.31} & \textbf{65.53}\\
\hline
\end{tabular}
\vspace{-1mm}
\caption{\textbf{Chain-of-thought reasoning results} on GSM8K and MMLU under MXFP4 format. Best results (excluding FP16) are in \textbf{bold}.}
\label{tab:instruct}
\vspace{-3mm}
\end{table}

\textbf{Zero-Shot Tasks.}
We further evaluate on five zero-shot benchmarks and report the average accuracy in Tables~\ref{tab:main_results} and~\ref{tab:full_results_llama}. FOCUS consistently achieves the highest average across nearly all model-format combinations. On Qwen3-4B, FOCUS obtains 67.05\% (NVFP4) and 65.65\% (MXFP4), recovering 98.2\% and 96.2\% of the FP16 accuracy (68.25\%), respectively. On Qwen3-8B NVFP4, FOCUS reaches 71.37\%, recovering 99.8\% of FP16 (71.54\%), while the best competing method RaZeR achieves only 70.54\%. Under MXFP4, FOCUS outperforms MR-GPTQ by 1.0\% in average accuracy.

\textbf{Reasoning Tasks.}
Table~\ref{tab:instruct} reports chain-of-thought evaluation results on GSM8K and MMLU under MXFP4. We observe that FP4 quantization incurs more pronounced degradation on reasoning tasks compared to perplexity and zero-shot benchmarks, indicating that multi-step reasoning is particularly sensitive to quantization noise. Nevertheless, FOCUS consistently preserves reasoning capability better than all competing methods. On LLaMA-3.1-8B-Instruct, FOCUS achieves 73.31\% on GSM8K and 65.53\% on MMLU, outperforming MR-GPTQ by 0.9 and 2.1 points, respectively. On LLaMA-3.2-3B-Instruct, FOCUS leads with 64.14\% (GSM8K) and 59.09\% (MMLU), surpassing MR-GPTQ by approximately 4 points on both benchmarks. These results show that FOCUS improves both token-level prediction and multi-step reasoning.

\begin{table}[t]
\centering
\begin{tabular}{l|l|ccc}
\hline
\textbf{Format} & \textbf{Method} & \textbf{Wiki2}$\downarrow$ & \textbf{C4}$\downarrow$ & \textbf{Avg.}$\uparrow$ \\
\hline
\multirow{4}{*}{NVFP4}
& RTN             & 13.88 & 17.29 & 65.58 \\
& Baseline        & 12.63 & 17.02 & 65.92 \\
& + CRS           & 12.61 & 16.98 & 66.91 \\
& + CRS + DGS     & \textbf{12.57} & \textbf{16.97} & \textbf{67.05} \\
\hline
\multirow{4}{*}{MXFP4}
& RTN             & 18.60 & 20.62 & 62.35 \\
& Baseline        & 12.96 & 17.98 & 64.69 \\
& + CRS           & 12.99 & 17.95 & 64.81 \\
& + CRS + DGS     & \textbf{12.85} & \textbf{17.77} & \textbf{65.65} \\
\hline
\end{tabular}
\caption{\textbf{Effect of CRS and DGS.} Each proposed component is progressively added to the naive scale learning baseline on Qwen3-4B. Best results are in \textbf{bold}.}
\label{tab:ablation}
\vspace{-5mm}
\end{table}

\subsection{Ablation Study}
\label{sec:ablation}
\textbf{Effect of CRS and DGS.}
Table~\ref{tab:ablation} presents an ablation study on Qwen3-4B, progressively adding each proposed component. Starting from RTN, the naive scale learning baseline already yields substantial gains, particularly under MXFP4 where Wiki2 perplexity drops from 18.60 to 12.96, confirming the value of end-to-end scale optimization. Adding CRS further improves zero-shot accuracy by up to 1.0\% under NVFP4, validating that relaxing the coupling between quantization and dequantization scales enables better FP4 assignments. Incorporating DGS provides further gains under both formats, showing that sub-block granularity refines FP4 assignments beyond block-level relaxation alone. The full method achieves the best results in all settings, improving zero-shot average by 1.13\% and 0.96\% over the baseline under NVFP4 and MXFP4, respectively.

\textbf{Effect of sub-block size in DGS.}
Table~\ref{tab:dgs_granularity} investigates the impact of sub-block granularity in DGS. When the sub-block size equals the native block size (32 for MXFP4, 16 for NVFP4), DGS reduces to CRS, as each block shares a single relaxation coefficient. Decreasing the sub-block size introduces finer-grained coefficients, enabling more precise control over rounding boundaries at the cost of additional trainable parameters during calibration. For MXFP4, reducing the sub-block size from 32 to 16 yields a notable improvement (e.g., Avg.: 64.81 $\to$ 65.65), while further reduction to 8 achieves comparable performance. For NVFP4, sub-block sizes 8 and 4 perform similarly, so we use size 8 for both formats to balance accuracy and cost.

\subsection{Quantization Cost}

As a PTQ method, FOCUS freezes the model weights and optimizes only scales and relaxation coefficients end-to-end, yielding a lightweight quantization process. We compare its cost with MR-GPTQ~\cite{mr-gptq}, the previous SOTA supporting both MXFP4 and NVFP4. As shown in Table~\ref{tab:quant_cost}, FOCUS quantizes faster, reducing time by 34\% on Qwen3-4B and 28\% on Qwen3-8B. MR-GPTQ requires several costly stages, including GPTQ-based reconstruction, online Hadamard transformation, activation reordering, and MSE-based scale search. In contrast, FOCUS jointly optimizes all scales with standard forward-backward passes, without iterative layer-wise processing. Although it uses more peak GPU memory for learnable coefficients and optimizer states, the overhead is moderate, and both methods run on a single NVIDIA H20 GPU.

\begin{table}[t]
\centering
\begin{tabular}{l|c|ccc}
\hline
\textbf{Format} & \textbf{Sub-block Size} & \textbf{Wiki2}$\downarrow$ & \textbf{C4}$\downarrow$ & \textbf{Avg.}$\uparrow$ \\
\hline
\multirow{4}{*}{NVFP4}
& 16 (CRS)  & 12.61 & 16.96 & 66.91 \\
& 8         & \textbf{12.57} & 16.97 & \textbf{67.05} \\
& 4         & \textbf{12.57} & \textbf{16.91} & \textbf{67.05} \\
& 2         & 12.62 & 16.92 & 66.94 \\
\hline
\multirow{4}{*}{MXFP4}
& 32 (CRS)  & 12.99 & 17.95 & 64.81 \\
& 16        & 12.98 & 17.79 & \textbf{65.65} \\
& 8         & 12.85 & \textbf{17.77} & \textbf{65.65} \\
& 4         & \textbf{12.83} & 17.78 & 64.94 \\
\hline
\end{tabular}
\caption{\textbf{Effect of sub-block size in DGS.} Evaluated on Qwen3-4B. Setting sub-block size to the native block size reduces DGS to CRS. Best results are in \textbf{bold}.}
\label{tab:dgs_granularity}
\vspace{-5mm}
\end{table}

\begin{table}[h]
\centering
\begin{tabular}{l|l|cc}
\hline
\textbf{Model} & \textbf{Method} & \textbf{GPU Memory} & \textbf{Time} \\
\hline
\multirow{2}{*}{Qwen3-4B} & MR-GPTQ & 45 GB & 29 min \\
& FOCUS & 77 GB & 19 min \\
\hline
\multirow{2}{*}{Qwen3-8B} & MR-GPTQ & 62 GB & 40 min \\
& FOCUS & 85 GB & 29 min \\
\hline
\end{tabular}
\caption{\textbf{Quantization cost} on one NVIDIA H20 GPU.}\label{tab:quant_cost}
\vspace{-5mm}
\end{table}

\section{Conclusion}
We present FOCUS, a post-training quantization framework that rethinks scale optimization for FP4 formats. Building on the observation that the quantization scale is free from hardware constraints, we construct an efficient end-to-end scale learning framework that exploits this overlooked freedom along two complementary dimensions. Coupled-Relaxation Scaling (CRS) lifts scale optimization from a discrete hardware-constrained space to a continuous one by introducing a learnable coefficient that relaxes the coupling between the quantization and dequantization scales. Dual-Granularity Scaling (DGS) further extends this idea to the sub-block level, enabling the relaxed scale to adapt to local weight variations within each block. Together, these techniques enable FOCUS to find superior FP4 assignments during calibration, while the deployed model remains in standard MXFP4/NVFP4 format with no additional inference cost. Experiments confirm that FOCUS achieves state-of-the-art FP4 accuracy under both formats, substantially closing the gap to full-precision performance.

\bigskip

\bibliography{aaai2027}

@article{46,
  title={Four Over Six: More Accurate NVFP4 Quantization with Adaptive Block Scaling},
  author={Cook, Jack and Guo, Junxian and Xiao, Guangxuan and Lin, Yujun and Han, Song},
  journal={arXiv preprint arXiv:2512.02010},
  year={2025}
}

@inproceedings{omniquant,
  title={OmniQuant: Omnidirectionally Calibrated Quantization for Large Language Models},
  author={Shao, Wenqi and Chen, Mengzhao and Zhang, Zhaoyang and Xu, Peng and Zhao, Lirui and Li, Zhiqian and Zhang, Kaipeng and Gao, Peng and Qiao, Yu and Luo, Ping},
  year={2023},
  booktitle={ICLR}
}

@inproceedings{lrquant,
  title={LRQuant: Learnable and Robust Post-Training Quantization for Large Language Models},
  author={Zhao, Jiaqi and Zhang, Miao and Zeng, Chao and Wang, Ming and Liu, Xuebo and Nie, Liqiang},
  year=2024,
  booktitle={ACL}
}

@inproceedings{spinquant,
  title={SpinQuant: LLM Quantization with Learned Rotations},
  author={Liu, Zechun and Zhao, Changsheng and Fedorov, Igor and Soran, Bilge and Choudhary, Dhruv and Krishnamoorthi, Raghuraman and Chandra, Vikas and Tian, Yuandong and Blankevoort, Tijmen},
  year=2025,
  booktitle={ICLR}
}

@inproceedings{affinequant,
  title={AffineQuant: Affine Transformation Quantization for Large Language Models},
  author={Ma, Yuexiao and Li, Huixia and Zheng, Xiawu and Ling, Feng and Xiao, Xuefeng and Wang, Rui and Wen, Shilei and Chao, Fei and Ji, Rongrong},
  year=2024,
  booktitle={ICLR}
}

@inproceedings{flatquant,
  title={FlatQuant: Flatness Matters for LLM Quantization},
  author={Sun, Yuxuan and Liu, Ruikang and Bai, Haoli and Bao, Han and Zhao, Kang and Li, Yuening and Hu, Jiaxin and Yu, Xianzhi and Hou, Lu and Yuan, Chun},
  year=2025,
  booktitle={ICML}
}

@inproceedings{adaround,
  title={Up or Down? Adaptive Rounding for Post-Training Quantization},
  author={Nagel, Markus and Amjad, Rana Ali and Van Baalen, Mart and Louizos, Christos and Blankevoort, Tijmen},
  year=2020,
  booktitle={ICML}
}

@inproceedings{brecq,
  title={BRECQ: Pushing the Limit of Post-Training Quantization by Block Reconstruction},
  author={Li, Yuhang and Gong, Ruihao and Tan, Xu and Yang, Yang and Hu, Peng and Zhang, Qi and Yu, Fengwei and Wang, Wei and Gu, Shi},
  year=2021,
  booktitle={ICLR}
}

@inproceedings{ostquant,
    title={OSTQuant: Refining Large Language Model Quantization with Orthogonal and Scaling Transformations for Better Distribution Fitting},
    author={Hu, Xing and Cheng, Yuan and Yang, Dawei and Chen, Zhixuan and Xu, Zukang and Yu, Jiangyong and Chen, Xu and Yuan, Zhihang and Jiang, Zhe and Zhou, Sifan},
    year=2025,
    booktitle={ICLR}
}

@inproceedings{mr-gptq,
    title={Bridging the Gap Between Promise and Performance for Microscaling FP4 Quantization},
    author={Egiazarian, Vage and Castro, Roberto L. and Kuznedelev, Denis and Panferov, Andrei and Kurtic, Eldar and Pandit, Shubhra and Marques, Alexandre Noll and Kurtz, Mark and Ashkboos, Saleh and Hoefler, Torsten and Alistarh, Dan},
    year=2026,
    booktitle={ICLR}
}

@inproceedings{blockrotation,
    title={Block Rotation is All You Need for MXFP4 Quantization},
    author={Shao, Yuantian and Wang, Peisong and Chen, Yuanteng and Xu, Chang and Wei, Zhihui and Cheng, Jian},
    year=2026,
    booktitle={ICML}
}

@inproceedings{arcquant,
    title={ARCQuant: Boosting NVFP4 Quantization with Augmented Residual Channels for LLMs},
    author={Meng, Haoqian and Luo, Yilun and Zhao, Yafei and Liu, Wenyuan and Zhang, Peng and Ma, Xindian},
    year=2026,
    booktitle={ACL}
}

@inproceedings{unveiling,
    title={Unveiling the Potential of Quantization with MXFP4: Strategies for Quantization Error Reduction},
    author={Chhugani, Jatin and Jeong, Geonhwa and Su, Bor-Yiing and Pan, Yunjie and Yang, Hanmei and Ankit, Aayush and Yu, Jiecao and Deng, Summer and Chen, Yunqing and Satish, Nadathur and Kim, Changkyu},
    year=2026,
    booktitle={ICML}
}

@article{batquant,
    title={BATQuant: Outlier-resilient MXFP4 Quantization via Learnable Block-wise Optimization},
    author={Li, Ji-Fu and Zhang, Manyi and Xia, Xiaobo and Bao, Han and Bai, Haoli and Dong, Zhenhua and Yu, Xianzhi},
    year=2026,
    journal={arXiv preprint arXiv:2603.16590}
}

@article{mxfp4,
  title={Microscaling Data Formats for Deep Learning},
  author={Rouhani, Bita Darvish and Zhao, Ritchie and More, Ankit and Hall, Mathew and Khodamoradi, Alireza and Deng, Summer and Choudhary, Dhruv and Cornea, Marius and Dellinger, Eric and Denolf, Kristof and others},
  journal={arXiv preprint arXiv:2310.10537},
  year={2023}
}

@article{nvfp4,
  title={Pretraining Large Language Models with NVFP4},
  author={Nvidia and Abecassis, Felix and Agrusa, Anjulie and Ahn, Dong and Alben, Jonah and Alborghetti, Stefania and Andersch, Michael and Arayandi, Sivakumar and Bjorlin, Alexis and Blakeman, Aaron and Briones, Evan and others},
  journal={arXiv preprint arXiv:2509.25149},
  year={2025}
}

@article{benchmarkingfp4,
  title={Benchmarking Post-Training Quantization of Large Language Models under Microscaling Floating Point Formats},
  author={Zhang, Manyi and Li, Ji-Fu and Sun, Zhongao and Bai, Haoli and Zhen, Hui-Ling and Dong, Zhenhua and Yu, Xianzhi},
  journal={arXiv preprint arXiv:2601.09555},
  year={2026}
}

@article{duquant++,
  title={DuQuant++: Fine-grained Rotation Enhances Microscaling FP4 Quantization},
  author={Lin, Haokun and Jia, Xinle and Xu, Haobo and Yao, Bingchen and Guo, Xianglong and Wu, Yichen and Lu, Zhichao and Wei, Ying and Zhang, Qingfu and Sun, Zhenan},
  journal={arXiv preprint arXiv:2604.17789},
  year={2026}
}

@article{faar,
  title={FAAR: Format-Aware Adaptive Rounding for NVFP4},
  author={Li, Hanglin and Tian, Shuchang and Lin, Chen and Zhao, Zhiyong and Zhan, Kun},
  journal={arXiv preprint arXiv:2603.22370},
  year={2026}
}

@inproceedings{gptq,
	title = {{GPTQ}: {Accurate} {Post}-{Training} {Quantization} for {Generative} {Pre}-trained {Transformers}},
	author = {Frantar, Elias and Ashkboos, Saleh and Hoefler, Torsten and Alistarh, Dan},
    booktitle ={ICLR},
	year = {2023},
}

@article{qwen3,
  title     = {Qwen3 Technical Report},
    author = {An Yang and Anfeng Li and Baosong Yang and Beichen Zhang and Binyuan Hui and Bo Zheng and Bowen Yu and Chang Gao and Chengen Huang and Chenxu Lv and others},
  journal   = {arXiv preprint arXiv:2505.09388},
  year      = {2025},
}

@article{llama3,
  title={The llama 3 herd of models},
  author={Dubey, Abhimanyu and Jauhri, Abhinav and Pandey, Abhinav and Kadian, Abhishek and Al-Dahle, Ahmad and Letman, Aiesha and Mathur, Akhil and Schelten, Alan and Yang, Amy and Fan, Angela and others},
  journal={arXiv preprint arXiv:2407.21783},
  year={2024}
}

@article{deepseek,
  title={Deepseek-r1: Incentivizing reasoning capability in llms via reinforcement learning},
  author={Guo, Daya and Yang, Dejian and Zhang, Haowei and Song, Junxiao and Zhang, Ruoyu and Xu, Runxin and Zhu, Qihao and Ma, Shirong and Wang, Peiyi and Bi, Xiao and others},
  journal={arXiv preprint arXiv:2501.12948},
  year={2025}
}

@article{kimi,
  title={Kimi k2: Open agentic intelligence},
  author={Team, Kimi and Bai, Yifan and Bao, Yiping and Charles, Y and Chen, Cheng and Chen, Guanduo and Chen, Haiting and Chen, Huarong and Chen, Jiahao and Chen, Ningxin and others},
  journal={arXiv preprint arXiv:2507.20534},
  year={2025}
}

@article{glm,
  title={Glm-5: from vibe coding to agentic engineering},
  author={Zeng, Aohan and Lv, Xin and Hou, Zhenyu and Du, Zhengxiao and Zheng, Qinkai and Chen, Bin and Yin, Da and Ge, Chendi and Huang, Chenghua and Xie, Chengxing and others},
  journal={arXiv preprint arXiv:2602.15763},
  year={2026}
}

@inproceedings{smoothquant,
	title = {{SmoothQuant}: {Accurate} and {Efficient} {Post}-{Training} {Quantization} for {Large} {Language} {Models}},
	author = {Xiao, Guangxuan and Lin, Ji and Seznec, Mickael and Wu, Hao and Demouth, Julien and Han, Song},
    booktitle={ICML},
	year = {2023},
}

@inproceedings{
    llmint8,
    title={{GPT}3.int8(): 8-bit Matrix Multiplication for Transformers at Scale},
    author={Tim Dettmers and Mike Lewis and Younes Belkada and Luke Zettlemoyer},
    booktitle={NeurIPS},
    year={2022},
}

@misc{blackwell,
  title={NVIDIA Blackwell Architecture Technical Brief},
  author={NVIDIA, Santa Clara},
  year={2024}
}

@inproceedings{quarot,
  title={Quarot: Outlier-free 4-bit inference in rotated llms},
  author={Ashkboos, Saleh and Mohtashami, Amirkeivan and Croci, Maximilian L and Li, Bo and Cameron, Pashmina and Jaggi, Martin and Alistarh, Dan and Hoefler, Torsten and Hensman, James},
  booktitle={NeurIPS},
  year={2024}
}

@article{razer,
  title={{RaZeR: Pushing the Limits of NVFP4 Quantization with Redundant Zero Remapping}},
  author={Yuzong Chen and Xilai Dai and Jake Hyun and Chi-Chih Chang and Wonsuk Jang and Yuheng Wu and Thierry Tambe and Jae-sun Seo and Mohamed S. Abdelfattah},
  journal={arXiv preprint arXiv:2501.04052v2},
  year={2026}
}

@inproceedings{pytorch,
  title={Pytorch: An imperative style, high-performance deep learning library},
  author={Paszke, Adam and Gross, Sam and Massa, Francisco and Lerer, Adam and Bradbury, James and Chanan, Gregory and Killeen, Trevor and Lin, Zeming and Gimelshein, Natalia and Antiga, Luca and others},
  booktitle={NeurIPS},
  year={2019}
}

@inproceedings{merity2016pointer,
  title={Pointer sentinel mixture models},
  author={Merity, Stephen and Xiong, Caiming and Bradbury, James and Socher, Richard},
  booktitle={ICLR},
  year={2017}
}

@article{raffel2020exploring,
  title={Exploring the limits of transfer learning with a unified text-to-text transformer},
  author={Raffel, Colin and Shazeer, Noam and Roberts, Adam and Lee, Katherine and Narang, Sharan and Matena, Michael and Zhou, Yanqi and Li, Wei and Liu, Peter J},
  journal={JMLR},
  year={2020}
}

@article{clark2018think,
  title={Think you have solved question answering? try arc, the ai2 reasoning challenge},
  author={Clark, Peter and Cowhey, Isaac and Etzioni, Oren and Khot, Tushar and Sabharwal, Ashish and Schoenick, Carissa and Tafjord, Oyvind},
  journal={arXiv preprint arXiv:1803.05457},
  year={2018}
}

@inproceedings{zellers2019hellaswag,
  title={Hellaswag: Can a machine really finish your sentence?},
  author={Zellers, Rowan and Holtzman, Ari and Bisk, Yonatan and Farhadi, Ali and Choi, Yejin},
  booktitle={ACL},
  year={2019}
}

@inproceedings{bisk2020piqa,
  title={Piqa: Reasoning about physical commonsense in natural language},
  author={Bisk, Yonatan and Zellers, Rowan and Gao, Jianfeng and Choi, Yejin and others},
  booktitle={AAAI},
  year={2020}
}

@inproceedings{sakaguchi2019adversarial,
  title={WINOGRANDE: An Adversarial Winograd Schema Challenge at Scale},
  author={Sakaguchi, Keisuke and Le Bras, Ronan and Bhagavatula, Chandra and Choi, Yejin},
  booktitle={AAAI},
  year={2020}
}

@inproceedings{adamw,
  title={Decoupled Weight Decay Regularization},
  author={Loshchilov, Ilya and Hutter, Frank},
  booktitle={ICLR},
  year={2019}
}

@inproceedings{llm-fp4,
  title={Llm-fp4: 4-bit floating-point quantized transformers},
  author={Liu, Shih-yang and Liu, Zechun and Huang, Xijie and Dong, Pingcheng and Cheng, Kwang-Ting},
  booktitle={EMNLP},
  year={2023}
}

@article{intvsfp,
  title={INT vs FP: A Comprehensive Study of Fine-Grained Low-bit Quantization Formats},
  author={Chen, Mengzhao and Wu, Meng and Jin, Hui and Yuan, Zhihang and Liu, Jing and Zhang, Chaoyi and Li, Yunshui and Huang, Jie and Ma, Jin and Xue, Zeyue and others},
  journal={arXiv preprint arXiv:2510.25602},
  year={2025}
}

@article{gsm8k,
  title={Training verifiers to solve math word problems},
  author={Cobbe, Karl and Kosaraju, Vineet and Bavarian, Mohammad and Chen, Mark and Jun, Heewoo and Kaiser, Lukasz and Plappert, Matthias and Tworek, Jerry and Hilton, Jacob and Nakano, Reiichiro and others},
  journal={arXiv preprint arXiv:2110.14168},
  year={2021}
}

@inproceedings{mmlu,
  title={Measuring massive multitask language understanding},
  author={Hendrycks, Dan and Burns, Collin and Basart, Steven and Zou, Andy and Mazeika, Mantas and Song, Dawn and Steinhardt, Jacob},
  booktitle={ICLR},
  year={2021}
}

@article{ste,
  title={Estimating or propagating gradients through stochastic neurons for conditional computation},
  author={Bengio, Yoshua and L{\'e}onard, Nicholas and Courville, Aaron},
  journal={arXiv preprint arXiv:1308.3432},
  year={2013}
}

@misc{mxfp4_ocp,
  title        = {{OCP Microscaling Formats (MX) Specification Version\,1.0}},
  author       = {{Open Compute Project Foundation (MX Alliance)}},
  howpublished = {Open Compute Project Foundation Technical Specification},
  year         = {2023},
  month        = sep,
  url          = {https://www.opencompute.org/documents/ocp-microscaling-formats-mx-v1-0-spec-final-pdf},
}

@article{soar,
  title={SOAR: Scale Optimization for Accurate Reconstruction in NVFP4 Quantization},
  author={Bao, Chengzhu and Yan, Xianglong and Li, Zhiteng and Qin, Guangshuo and Yu, Guanghua and Zhang, Yulun},
  journal={arXiv preprint arXiv:2605.12245},
  year={2026}
}

@inproceedings{pt2-llm,
  title={PT$^2$-LLM: Post-Training Ternarization for Large Language Models},
  author={Yan, Xianglong and Bao, Chengzhu and Li, Zhiteng and Zhang, Tianao and Yang, Kaicheng and Qin, Haotong and Xie, Ruobing and Sun, Xingwu and Zhang, Yulun},
  booktitle={ICLR},
  year={2026}
}

@inproceedings{arb-llm,
  title={Arb-llm: Alternating refined binarizations for large language models},
  author={Li, Zhiteng and Yan, Xianglong and Zhang, Tianao and Qin, Haotong and Xie, Dong and Tian, Jiang and Kong, Linghe and Zhang, Yulun and Yang, Xiaokang and others},
  booktitle={ICLR},
  year={2025}
}

@inproceedings{huggingface,
  title={Transformers: State-of-the-art natural language processing},
  author={Wolf, Thomas and Debut, Lysandre and Sanh, Victor and Chaumond, Julien and Delangue, Clement and Moi, Anthony and Cistac, Pierric and Rault, Tim and Louf, R{\'e}mi and Funtowicz, Morgan and others},
  booktitle={EMNLP},
  year={2020}
}


\end{document}